\pdfoutput=1
\documentclass{bmvc2k}

\title{MedPlex: Deep Vision-Language Co-Adaptation for Clinically Grounded Medical Segmentation}

\runninghead{Sultan et al.}{MedPlex: Vision-Language Co-Adaptation}

\addauthor{Rafi Ibn Sultan}{rafis@wayne.edu}{1}
\addauthor{Hui Zhu}{hui@wayne.edu}{1}
\addauthor{Chengyin Li}{cli6@hfhs.org}{2}
\addauthor{Dongxiao Zhu}{dzhu@wayne.edu}{1,3}

\addinstitution{
Department of Computer Science\\
Wayne State University\\
}
\addinstitution{
 Department of Radiation Oncology\\
 Henry Ford Health
}
\addinstitution{
 Institute for AI and Data Science\\
 Wayne State University
}

\begin{document}

\maketitle

\begin{abstract}
Medical image segmentation is still largely trained as a vision-only problem, even
though clinical interpretation often relies on textual knowledge about anatomy,
location, appearance, and surrounding context. Text-guided segmentation methods
within the Vision-Language Model (VLM) paradigm seek to bring this complementary
knowledge into dense prediction, but language is often used only as a late
conditioning signal: encoded once, introduced after substantial visual abstraction,
and weakly connected to how visual representations are formed. We argue that text
can meaningfully support segmentation only when it remains active throughout visual
representation learning and is organized around the clinical concepts that
distinguish anatomical structures. We introduce \textbf{MedPlex}
(\textbf{Med}ical \textbf{Plex}us of Vision and Language), an end-to-end VLM
framework that turns text guidance into a continuous, clinically grounded learning
process. Through \textbf{BiFusion} (\textbf{Bi}directional \textbf{Fusion}),
visual and textual representations evolve together across the encoding hierarchy,
while \emph{class-level concept alignment} and \emph{region-level concept alignment}
structure the shared representation at two complementary granularities. Class-level
concept alignment anchors each anatomical target to its aggregated clinical concept
profile, while region-level concept alignment preserves the contribution of
individual clinical concepts (e.g., shape, location, appearance, and texture)
through class-specific visual evidence. In this way, MedPlex uses language as structured supervision for learning segmentation representations throughout the encoder. MedPlex achieves state-of-the-art performance across CT
and MR benchmarks covering multi-organ, cardiac substructure, and tumor
segmentation, including settings with real free-text clinical supervision. The source code is available in our \href{https://github.com/rafiibnsultan/MedPlex}{GitHub repository}. 
\end{abstract}

\section{Introduction}
\label{sec:intro}

Medical image segmentation has advanced rapidly through architectures trained directly on image-mask supervision, with purely visual models achieving strong performance across diverse anatomical targets and imaging modalities~\cite{ronneberger2015unet,hatamizadeh2021swin,li2023focalunetr,li2023autoprosam,sultan2026neighborhoodattentiontransformernetwork, li2023new, mgboh2025fluenceformer}. Yet this dominant formulation treats segmentation as a visual recognition problem alone. In clinical practice, scans are interpreted together with anatomical knowledge, structured reports, clinical notes, and textual descriptions that specify what a structure is, where it is expected to appear, how it relates to surrounding tissue, and how its appearance may vary across patients and imaging protocols~\cite{oh2023llm,zhao2023one,zhao2025foundation}. Such textual cues provide compact semantic and relational knowledge that image-mask supervision does not explicitly encode.

This has motivated text-guided segmentation methods within the Vision-Language Model (VLM) paradigm, which use textual descriptions to guide dense visual prediction~\cite{radford2021learning,luddecke2022image,huang2024cat,liu2023clip}. The premise is compelling: language should help disambiguate visually similar regions, inject anatomical context, and support generalization beyond fixed label vocabularies. However, existing text-guided methods often struggle to make language a meaningful part of representation learning. This suggests that the bottleneck lies not in whether language is useful, but in how it is represented and integrated with vision.

\begin{wrapfigure}{r}{0.6\columnwidth}
\includegraphics[width=0.6\columnwidth]{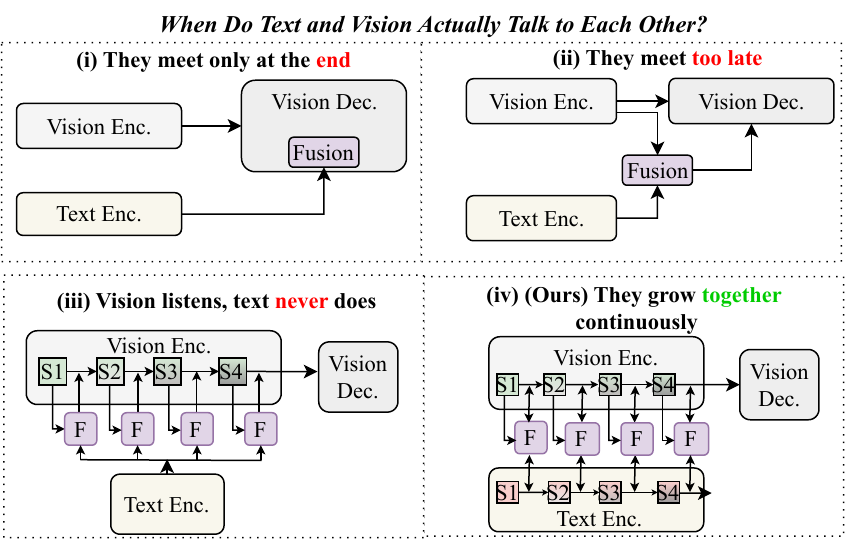}
\caption{\small How and when text and vision interact determines whether textual knowledge can meaningfully guide segmentation. Our approach (iv) lets both modalities continuously inform each other throughout processing, while prior approaches (i--iii) either meet too late or only listen in one direction.}
\label{fig:main_idea}
\end{wrapfigure}

We argue that a central limitation of existing medical VLM segmentation models is \emph{representation staleness}. As illustrated in \Cref{fig:main_idea}, most approaches first build a visual hierarchy and then inject text after substantial visual abstraction has already occurred, either at the end of the visual pipeline~\cite{zhong2023ariadne,zeng2024abp,guo2024common,zhang2024madapter}, near the top of the visual hierarchy~\cite{zhao2023one,liu2023clip,jiang2024zept,wang2022cris}, or by broadcasting a fixed text representation across visual stages~\cite{feng2021encoder,yang2022lavt,oh2023llm,li2023lvit}. These designs let language modulate visual features, but the interaction is delayed or asymmetric: visual features are formed for many layers without linguistic context, while textual representations are usually computed once and remain blind to the image they are meant to guide. As a result, text can act as a late conditioning signal, but it cannot continuously shape how anatomical evidence is built from local appearance to higher-level semantics. This is especially limiting because vision and language do not become useful only at their endpoints. Visual encoders progressively transform local intensity patterns into semantic anatomical representations~\cite{dosovitskiy2020image,hatamizadeh2021swin}, while language encoders move from lexical and syntactic cues toward global meaning~\cite{tenney2019bert,peters2018dissecting}. Existing designs therefore miss the opportunity to align the two modalities throughout their parallel abstraction processes.

A second bottleneck is that the textual signal itself is often too coarse for dense medical prediction. Medical language is structured, compositional, and domain-specific: clinically meaningful descriptions often depend on subtle qualifiers of shape, position, appearance, texture, and anatomical context. General-purpose text encoders are not optimized to preserve these distinctions in a form that is directly useful for segmentation~\cite{huang2024cat,oh2023llm}. Consequently, the text embedding supplied to a segmentation model may fail to distinguish closely related anatomical classes or collapse descriptions that differ in clinically important ways. Even a strong fusion module cannot extract reliable guidance from a representation that does not preserve the concepts needed for localization and boundary delineation. Thus, text-guided segmentation faces two coupled problems: language must interact with vision while both representations are being constructed, and textual representations must be organized around clinically meaningful concepts instead of being treated as a late conditioning signal.

We introduce \textbf{MedPlex} (\textbf{Med}ical \textbf{Plex}us of Vision and Language), an end-to-end VLM framework that turns text guidance into a continuous, clinically grounded learning process. MedPlex realizes this through \textbf{BiFusion} (\textbf{Bi}directional \textbf{Fusion}), which progressively updates both visual and textual encoder streams across the encoding hierarchy. At each stage, visual features query language for semantic guidance, while textual representations are grounded using image-specific visual evidence. This keeps both modalities adaptive, allowing language to shape how visual evidence is organized and allowing textual representations to evolve with the current image. To structure this interaction around clinically meaningful knowledge, MedPlex further introduces \emph{class-level concept alignment} and \emph{region-level concept alignment}. MedPlex organizes supervision around clinical concepts such as shape, location, appearance, and texture. Class-level concept alignment anchors anatomical classes to aggregated clinical concept descriptions, while region-level concept alignment preserves the contribution of individual clinical concepts through class-specific visual evidence. Together, these alignments connect language to both the semantic identity of each anatomical target and the fine-grained clinical concepts that support its segmentation.

Our contributions are summarized as follows:
\begin{itemize}
\item We introduce \textbf{MedPlex}, a unified end-to-end framework for text-guided medical image segmentation within the VLM paradigm, compatible with standard encoder-decoder segmentation backbones and applicable to both structured textual descriptions and real free-text clinical supervision.

\item We develop three reusable components for clinically grounded vision-language segmentation: \textbf{BiFusion} (\textbf{Bi}directional \textbf{Fusion}), \emph{class-level concept alignment}, and \emph{region-level concept alignment}.

\item We validate MedPlex across diverse CT and MR benchmarks spanning multi-organ, cardiac substructure, and tumor segmentation, where it achieves state-of-the-art performance and demonstrates robust performance across modalities, anatomical targets, and textual supervision regimes.
\end{itemize}

\section{Related Work}
\label{sec:related_work}

\subsection{Vision-Language Fusion Architectures}

Language integration has been widely studied for segmentation, where textual descriptions help specify what should be segmented, disambiguate visually similar regions, and support open-vocabulary or referring segmentation~\cite{liang2023open,xu2022simple,luddecke2022image,li2024deal,huang2024cross}. In 3D medical image segmentation, however, vision-language interaction remains less developed. Existing methods differ mainly in how textual descriptions enter the segmentation pipeline, which visual representations they influence, and whether textual representations can adapt to image-specific visual evidence.

\noindent\textbf{Fusion at Vision Decoder.}
A common strategy is to process image and text independently, then fuse them near the decoder or final prediction stage~\cite{zeng2024abp,shastri2024locate,huemann2024contextual,li2024textmatch,du2024segvol,zeng2025harnessing}. This design is simple and compatible with existing encoder-decoder segmentation models, but it makes language primarily a late conditioning signal. Since the visual hierarchy has already been formed, textual descriptions can influence decoding but have limited ability to shape the visual representations that feed it.

\noindent\textbf{Late Fusion at Vision Encoder.}
Other approaches move interaction earlier by injecting language near the top of the visual encoder. These methods use either short text phrases~\cite{liu2023clip,li2024novel} or detailed class prompts~\cite{zhao2023one,zhou2025sam,jiang2024zept,huang2024cat,shui2025largescale,rokuss2025voxtell,yuan2025tgsam,zeng2026decoupling} to modulate high-level visual features. Although this gives language access to encoder representations before decoding, the two modalities still develop largely in isolation through most of the visual hierarchy. As a result, language can refine high-level visual abstraction, but it remains weakly involved in how visual evidence is progressively constructed.

\noindent\textbf{Unidirectional Fusion at Vision Encoder.}
A broader line of work distributes textual signals across multiple visual processing stages~\cite{li2023lvit,hu2024lga,bui2024visual,zhong2023ariadne,guo2024common,oh2023llm,kim2023ro}. These approaches improve coverage of the visual hierarchy, but most interactions remain unidirectional: textual representations guide visual features, while visual evidence does not update the textual representations in return. This leaves language image-agnostic even when it is repeatedly injected into the visual stream. MedPlex differs through BiFusion, which directly updates both visual and textual encoder streams through bidirectional interaction across the hierarchy, allowing their representations to co-adapt as they evolve.

\subsection{Vision-Language Embedding Alignment}

Effective fusion depends on textual representations that preserve the clinical concepts needed for dense prediction. Biomedical and clinical text encoders~\cite{lee2020biobert,alsentzer2019publicly,gu2021domain} provide useful domain priors, while global contrastive learning has been used to align image and text representations~\cite{chen2023generative,jiang2024zept,ding2024cat,zhao2023one,hamamci2024developing}. However, anatomical targets often share overlapping clinical concepts, such as shape, location, appearance, and texture, making a single global text embedding insufficient for precise segmentation guidance.

More structured alignment has been explored in 2D chest X-ray pretraining by aligning image patches with report tokens~\cite{huang2021gloria,wang2022multi,muller2022joint} or predefined phrase groups~\cite{liang2025medfilip,liu2023improving,phan2024decomposing}. In 3D medical imaging, alignment is often inferred post hoc from segmentation masks~\cite{shui2025largescale} or applied mainly to decoder-level features~\cite{jiang2024unleashing}, leaving encoder representations less directly organized around fine-grained clinical concepts. MedPlex instead applies \emph{class-level concept alignment} and \emph{region-level concept alignment} immediately after bidirectional encoder fusion, connecting clinical concepts to anatomical classes and class-specific visual evidence.

\begin{figure*}[t]
\centering
\includegraphics[width=1.0\textwidth]{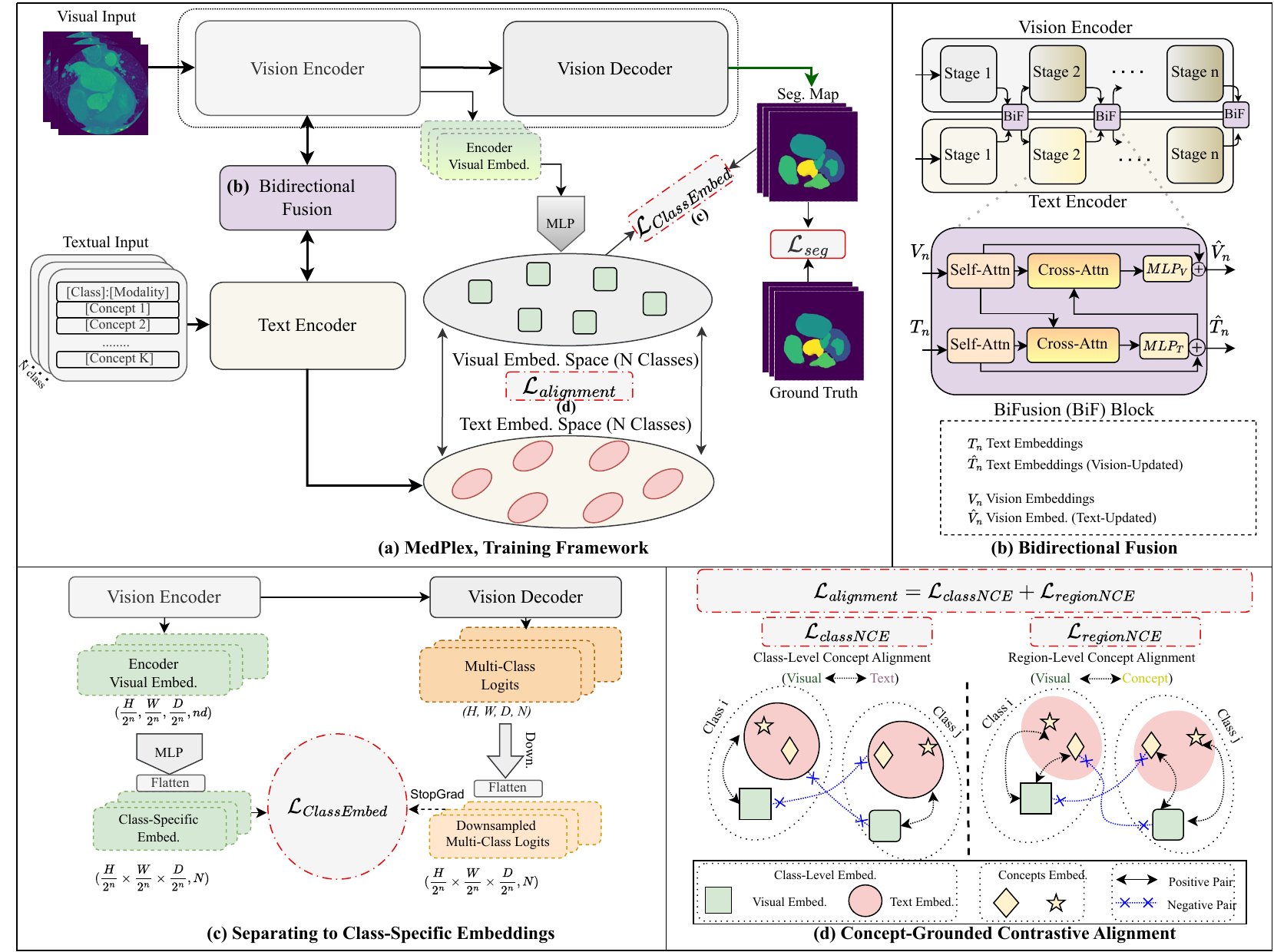}
\caption{\small
(a) \textbf{MedPlex}, an end-to-end training pipeline that integrates vision and language for medical image segmentation.
(b) \textbf{Bidirectional fusion} between visual and text encoders through BiFusion blocks at each stage.
(c) $\mathcal{L}_{\mathrm{ClassEmbed}}$: mapping vision encoder embeddings to class-specific embeddings with decoder supervision.
(d) $\mathcal{L}_{\mathrm{alignment}}$: concept-grounded contrastive alignment operating at two granularities. Class-level concept alignment aggregates concept embeddings into a single textual representation matched to class-wise visual features, while region-level concept alignment aligns individual concept embeddings with class-specific visual evidence.
}
\label{fig:figure2}
\end{figure*}

\section{Method}
\label{sec:method}

MedPlex integrates textual descriptions directly into the formation of segmentation representations throughout the encoder. As shown in \Cref{fig:figure2}a, MedPlex jointly processes a 3D medical image and class-specific textual descriptions in an end-to-end framework. The model first progressively updates hierarchical visual and textual encoder representations through BiFusion (\Cref{fig:figure2}b), then converts encoder visual representations into class-structured responses for alignment-ready supervision (\Cref{fig:figure2}c), and finally applies concept-grounded alignment at both class and region levels (\Cref{fig:figure2}d). The full framework is optimized with segmentation, class-response, and alignment objectives.

\subsection{Problem Setup}
\label{sec:overview}

Let $\mathbf{X} \in \mathbb{R}^{H \times W \times D}$ denote a 3D medical image and $\mathbf{G} \in \{0,1\}^{H \times W \times D \times N}$ its corresponding segmentation mask over $N$ anatomical classes. MedPlex receives textual supervision in two forms: real free-text clinical reports when report-level supervision is available, and structured class-specific textual descriptions when reports are unavailable. The model predicts a segmentation map $\mathbf{P}$ and jointly learns visual and textual representations through encoder-level fusion and concept-grounded alignment.

\subsection{Input Design and Modality Encoding}
\label{sec:input_encoding}

MedPlex integrates a standard 3D encoder-decoder segmentation backbone with a transformer-based text encoder. The visual stream processes 3D medical images and produces multi-stage visual representations, while the language stream processes textual descriptions and produces hierarchical textual representations. The two streams are matched stage by stage so BiFusion can update visual and textual representations at corresponding levels of abstraction.

\begin{wrapfigure}{r}{0.65\textwidth}
\includegraphics[width=0.95\linewidth]{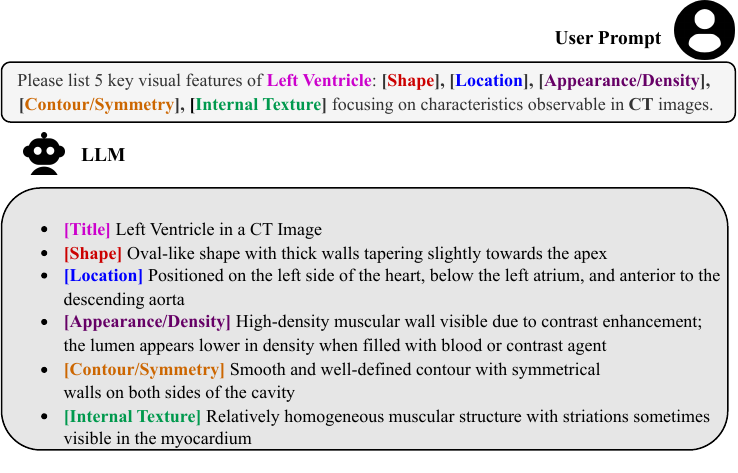}
\caption{\small A \textit{simplified demo} of curating structured concept descriptions of each class in our datasets using an LLM.}
\label{fig:figure3}
\end{wrapfigure}

\noindent\textbf{Textual Input.}
MedPlex supports two textual supervision formats: free-text clinical reports and structured class-specific textual descriptions. For datasets without reports, we construct class-specific textual descriptions using clinically observable concepts: shape, location, appearance/density, contour/symmetry, and internal texture in the target modality. Each concept provides a separate textual anchor for concept-grounded alignment, allowing MedPlex to preserve concept-level information instead of compressing the class description into a single text embedding. The descriptions are generated once using an LLM (\Cref{fig:figure3}), manually reviewed, and fixed during training. Dataset-specific details and prompt templates are provided in the Appendix.

\noindent\textbf{Text Encoder.}
MedPlex uses a transformer text encoder to produce stage-wise textual representations for BiFusion. Standard transformer encoders~\cite{devlin2018bert,vaswani2017attention} process text as a sequence, whereas visual encoders form a hierarchy through progressive resolution changes. To make the two streams compatible across stages, we partition a BERT encoder~\cite{devlin2018bert} into four groups following~\cite{cho2023cross}: layers [1--6], [7--8], [9--10], and [11--12].
These groups 
provide textual representations at progressively higher levels of abstraction~\cite{tenney2019bert}. Early BiFusion stages therefore interact with lower-level textual representations, while later stages interact with more semantic textual representations. This stage-wise design avoids forcing every visual stage to use the same final-layer text encoder output. The text encoder uses a feature dimension of 768 and can be initialized from domain-specific BERT checkpoints~\cite{lee2020biobert,alsentzer2019publicly,gu2021domain}.

\noindent\textbf{Vision Model.}
The visual stream uses a standard encoder-decoder 3D segmentation architecture. The encoder progressively extracts visual representations at multiple resolutions, and the decoder predicts the final segmentation map using the encoded features and skip connections. The segmentation objective combines Dice and Cross-Entropy losses:
\begin{equation}
\mathcal{L}_{\mathrm{seg}} =
\mathcal{L}_{\mathrm{dice}}(\mathbf{P}, \mathbf{G})
+
\mathcal{L}_{\mathrm{ce}}(\mathbf{P}, \mathbf{G}),
\label{equ:task1}
\end{equation}
where $\mathbf{P}$ denotes the predicted segmentation map and $\mathbf{G}$ denotes the ground truth mask.

\subsection{BiFusion: Continuous Vision-Language Co-adaptation}
\label{sec:bifusion}

BiFusion keeps visual and textual representations connected while both streams evolve through the encoder. At stage $n$, the visual stream provides image-specific visual evidence, while the language stream provides semantic context from the textual descriptions. BiFusion updates both streams at each stage: textual representations first absorb image-specific visual evidence, and visual features are then updated using the resulting image-conditioned textual context.

Let $\mathbf{T}_n \in \mathbb{R}^{N_t \times C_t}$ denote the textual representations at encoder stage $n$, where $N_t$ is the number of text tokens and $C_t=768$ for BERT. Let $\mathbf{V}_n \in \mathbb{R}^{H_n \times W_n \times D_n \times C_n}$ denote the visual feature map at the same stage. We flatten $\mathbf{V}_n$ into spatial tokens before attention and apply stage-specific projections so that visual and textual representations share a common attention dimension. BiFusion can then be viewed as a stage-wise co-adaptation operator,
\begin{equation}
(\hat{\mathbf{T}}_n,\hat{\mathbf{V}}_n)
=
\mathrm{BiF}_n(\mathbf{T}_n,\mathbf{V}_n),
\label{equ:bif_operator}
\end{equation}
which first builds modality-specific contextual representations and then performs two coupled cross-modal updates:
\begin{equation}
\begin{aligned}
\mathbf{T}'_n &= \mathrm{SA}\!\left(\mathrm{LN}(\mathbf{T}_n)\right),
\qquad
\mathbf{V}'_n = \mathrm{SA}\!\left(\mathrm{LN}(\mathbf{V}_n)\right), \\
\Delta\mathbf{T}_n
&=
\mathrm{MLP}_T\!\left(
\mathrm{CA}(\mathbf{T}'_n,\mathbf{V}'_n)
\right), \qquad
\hat{\mathbf{T}}_n
=
\mathbf{T}'_n + \Delta\mathbf{T}_n, \\
\Delta\mathbf{V}_n
&=
\mathrm{MLP}_V\!\left(
\mathrm{CA}(\mathbf{V}'_n,\hat{\mathbf{T}}_n)
\right), \qquad
\hat{\mathbf{V}}_n
=
\mathbf{V}'_n + \Delta\mathbf{V}_n .
\end{aligned}
\label{equ:bifusion}
\end{equation}
Here, $\Delta\mathbf{T}_n$ denotes the image-conditioned update to the textual representations, and $\Delta\mathbf{V}_n$ denotes the text-conditioned update to the visual representations. The first cross-attention step grounds textual representations in image-specific visual evidence, while the second returns the updated textual context to the visual stream. The resulting pair $(\hat{\mathbf{T}}_n,\hat{\mathbf{V}}_n)$ is passed to the next encoder stage, enabling co-adaptation to accumulate across the hierarchy.

\subsection{Class-Structured Encoder Representations}
\label{sec:class_structured_encoder}

The BiFusion-updated encoder feature map $\hat{\mathbf{V}}_n$ contains visual embeddings from all anatomical classes. In a standard segmentation model, these embeddings are separated into class-specific predictions only later by the decoder head. However, MedPlex applies concept-grounded alignment before the final decoder output is produced. We therefore need an intermediate representation that separates encoder features by anatomical class, allowing visual evidence to be compared with class-level and concept-level textual representations. To obtain this representation, MedPlex introduces a lightweight projection $\mathrm{MLP}_{\mathrm{emb}}$ that maps each spatial location in $\hat{\mathbf{V}}_n$ into an $N$-dimensional class-response vector:
\begin{equation}
\mathbf{V}_{\mathrm{cls}}
=
\phi_{\mathrm{cls}}(\hat{\mathbf{V}}_n)
=
\mathrm{MLP}_{\mathrm{emb}}(\hat{\mathbf{V}}_n)
\in
\mathbb{R}^{H_n \times W_n \times D_n \times N},
\label{equ:class_response}
\end{equation}
where $N$ is the number of anatomical classes and $\phi_{\mathrm{cls}}$ denotes the encoder-level class-response projection. Each channel of $\mathbf{V}_{\mathrm{cls}}$ corresponds to one anatomical class and provides the class-specific encoder representation used for concept alignment.

The remaining question is how to supervise this projection, since no separate class labels are available at the encoder stage. We therefore use the model's own decoder prediction $\mathbf{P}$ as a supervision signal. The decoder remains responsible for producing the final segmentation map, while the class-response projection learns to expose a class-structured view of the encoder representation. Specifically, we use trilinear interpolation to downsample $\mathbf{P}$ to the encoder resolution and obtain a soft target for $\mathbf{V}_{\mathrm{cls}}$:
\begin{equation}
\tilde{\mathbf{P}}_n
=
\mathrm{Down}(\mathbf{P})
\in
\mathbb{R}^{H_n \times W_n \times D_n \times N}.
\end{equation}
Before computing the class-response loss, we detach the decoder-derived target $\tilde{\mathbf{P}}_n$ from the computation graph, ensuring that this auxiliary objective does not propagate gradients into the decoder. The projected class-response map is then trained to match the detached target (where $\operatorname{sg}(\cdot)$ denotes the stop-gradient operation):
\begin{equation}
\mathcal{L}_{\mathrm{ClassEmbed}}
=
\mathrm{CE}
\left(
\mathbf{V}_{\mathrm{cls}},
\operatorname{sg}\left(\tilde{\mathbf{P}}_n\right)
\right).
\label{equ:class_embed}
\end{equation}
This objective encourages encoder features to expose class-discriminative responses before the decoder stage, giving class-level and region-level concept alignment the class-specific visual evidence needed to compare with textual concepts. For subsequent concept alignment, the class-wise visual embedding $\mathbf{v}^{(i)}$ is obtained by average-pooling the $i$-th class channel of $\mathbf{V}_{\mathrm{cls}}$ over the spatial dimensions.

\subsection{Concept-Grounded Alignment}
\label{sec:concept_alignment}

After BiFusion and class-structured projection, MedPlex applies concept-grounded alignment at two levels: \emph{class-level concept alignment} for anatomical-class semantics and \emph{region-level concept alignment} for concept-wise visual evidence.

\subsubsection{Class-Level Concept Alignment}
\label{sec:class_alignment}

For each anatomical class $i$, the textual description is decomposed into $K$ clinical concepts, such as shape, location, appearance, and texture. Let $\{\mathbf{t}^{(i)}_k\}_{k=1}^{K}$ denote the corresponding concept embeddings for class $i$. Given the class-structured encoder representation $\mathbf{V}_{\mathrm{cls}}$, we obtain a class-wise visual embedding $\mathbf{v}^{(i)}$ for each anatomical class. MedPlex first aggregates the concept embeddings into an image-conditioned textual representation for each anatomical class. Both visual and textual concept embeddings are projected into a shared normalized space, and concept relevance is computed from their cosine similarity:
\begin{equation}
\begin{aligned}
\mathbf{t}'^{(i)}_k &= \mathrm{norm}\!\left(\mathrm{Proj}_t(\mathbf{t}^{(i)}_k)\right),
\qquad
\mathbf{v}'^{(i)} = \mathrm{norm}\!\left(\mathrm{Proj}_v(\mathbf{v}^{(i)})\right), \\
w^{(i)}_k &=
\frac{\exp\!\left(\cos(\mathbf{t}'^{(i)}_k,\mathbf{v}'^{(i)})\right)}
{\sum_{\ell=1}^{K}\exp\!\left(\cos(\mathbf{t}'^{(i)}_\ell,\mathbf{v}'^{(i)})\right)},
\qquad
\mathbf{T}^{(i)} = \sum_{k=1}^{K} w^{(i)}_k \mathbf{t}^{(i)}_k .
\end{aligned}
\label{equ:concept_aggregation}
\end{equation}
The weights $w^{(i)}_k$ allow each class representation to emphasize the clinical concepts most relevant to the input image, producing an image-conditioned textual representation $\mathbf{T}^{(i)}$ while preserving the concept structure of the original textual description.

We then align class-wise visual embeddings and aggregated textual representations across the $N$ anatomical classes using bidirectional InfoNCE. To prevent premature convergence and reduce overconfident early alignment, small Gaussian noise $\mathcal{N}(0,\sigma^2)$ is added to both modalities during contrastive learning. For the visual-to-textual direction, the similarity matrix and contrastive loss are:
\begin{equation}
\begin{aligned}
\mathbf{S}_{ij}^{\mathbf{V}\rightarrow\mathbf{T}}
&=
\frac{\mathbf{v}^{(i)} \cdot \mathbf{T}^{(j)}}
{\|\mathbf{v}^{(i)}\|_2 \|\mathbf{T}^{(j)}\|_2}, \\
\mathcal{L}_{\mathrm{InfoNCE}}^{\mathbf{V}\rightarrow\mathbf{T}}
&=
-\frac{1}{N}\sum_{i=1}^{N}
\log
\frac{
\exp\!\left(\mathbf{S}_{ii}^{\mathbf{V}\rightarrow\mathbf{T}}/\tau\right)
}{
\sum_{j=1}^{N}
\exp\!\left(\mathbf{S}_{ij}^{\mathbf{V}\rightarrow\mathbf{T}}/\tau\right)
}.
\end{aligned}
\label{equ:class_nce_vt}
\end{equation}

The textual-to-visual direction $\mathcal{L}_{\mathrm{InfoNCE}}^{\mathbf{T}\rightarrow\mathbf{V}}$ is defined symmetrically. The resulting bidirectional objective aligns each anatomical class with its clinical concept profile while separating it from other anatomical classes:
\begin{equation}
\begin{aligned}
\mathcal{L}_{\mathrm{classNCE}}
&=
\frac{1}{2}
\left(
\mathcal{L}_{\mathrm{InfoNCE}}^{\mathbf{V}\rightarrow\mathbf{T}}
+
\mathcal{L}_{\mathrm{InfoNCE}}^{\mathbf{T}\rightarrow\mathbf{V}}
\right).
\end{aligned}
\label{equ:class_nce}
\end{equation}

\subsubsection{Region-Level Concept Alignment}
\label{sec:region_alignment}

Class-level concept alignment aggregates the clinical concepts of each anatomical class into a single textual representation. Region-level concept alignment provides finer supervision by aligning each individual clinical concept with the corresponding class-specific visual embedding. For class $i$ and concept $k$, MedPlex compares the concept embedding $\mathbf{t}^{(i)}_k$ with the class-wise visual embedding $\mathbf{v}^{(i)}$ derived from the class-structured encoder representation. This preserves the contribution of individual clinical concepts, such as shape, location, appearance, and texture, while keeping the alignment tied to the corresponding anatomical class.

Because different anatomical classes can share similar clinical concepts, using all mismatched classes as negatives can introduce weak or noisy supervision. MedPlex therefore restricts the contrastive objective to a hard-negative set. For each concept $k$, we compute concept-to-visual similarities across anatomical classes, select the top $\mathcal{K}$ mismatched class embeddings as negatives, and optimize the concept-to-visual direction as follows:
\begin{equation}
\begin{aligned}
\mathbf{S}_{ij}^{\mathbf{t}_k \rightarrow \mathbf{V}}
&=
\frac{
\mathbf{t}^{(i)}_k \cdot \mathbf{v}^{(j)}
}{
\|\mathbf{t}^{(i)}_k\|_2 \|\mathbf{v}^{(j)}\|_2
},
\qquad
\mathcal{N}_{\mathcal{K}}(i,k)
=
\operatorname{TopK}_{j \neq i}^{\mathcal{K}}
\left(
\mathbf{S}_{ij}^{\mathbf{t}_k \rightarrow \mathbf{V}}
\right), \\
\mathcal{L}_{\mathrm{InfoNCE}}^{\mathbf{t}_k \rightarrow \mathbf{V}}
&=
-\frac{1}{N}\sum_{i=1}^{N}
\log
\frac{
\exp\!\left(\mathbf{S}_{ii}^{\mathbf{t}_k \rightarrow \mathbf{V}}/\tau\right)
}{
\exp\!\left(\mathbf{S}_{ii}^{\mathbf{t}_k \rightarrow \mathbf{V}}/\tau\right)
+
\sum_{j\in\mathcal{N}_{\mathcal{K}}(i,k)}
\exp\!\left(\mathbf{S}_{ij}^{\mathbf{t}_k \rightarrow \mathbf{V}}/\tau\right)
}.
\end{aligned}
\label{equ:region_nce_tv}
\end{equation}

The visual-to-concept direction $\mathcal{L}_{\mathrm{InfoNCE}}^{\mathbf{V} \rightarrow \mathbf{t}_k}$ is defined symmetrically. Averaging over all $K$ clinical concepts gives the region-level concept alignment loss, and the full concept-grounded alignment objective combines the class-level and region-level terms using $\alpha_1$ and $\alpha_2$ to control their relative contributions:
\begin{equation}
\begin{aligned}
\mathcal{L}_{\mathrm{regionNCE}}
&=
\frac{1}{2K}
\sum_{k=1}^{K}
\left(
\mathcal{L}_{\mathrm{InfoNCE}}^{\mathbf{V} \rightarrow \mathbf{t}_k}
+
\mathcal{L}_{\mathrm{InfoNCE}}^{\mathbf{t}_k \rightarrow \mathbf{V}}
\right), \\
\mathcal{L}_{\mathrm{alignment}}
&=
\alpha_1 \mathcal{L}_{\mathrm{classNCE}}
+
\alpha_2 \mathcal{L}_{\mathrm{regionNCE}} .
\end{aligned}
\label{equ:region_alignment_full}
\end{equation}

\subsection{Training Objective}
\label{sec:training_objective}

MedPlex is trained end-to-end with a joint objective over segmentation, class-response supervision, and concept-grounded alignment. The segmentation loss trains the decoder to predict the final mask, the class-response loss makes encoder representations alignment-ready, and the alignment loss organizes the shared vision-language space around clinical concepts. The full training objective is:
\begin{equation}
\mathcal{L}
=
\beta_1 \mathcal{L}_{\mathrm{seg}}
+
\beta_2 \mathcal{L}_{\mathrm{ClassEmbed}}
+
\beta_3 \mathcal{L}_{\mathrm{alignment}} .
\label{equ:overall_loss}
\end{equation}
Following homoscedastic uncertainty weighting~\cite{kendall2018multi}, $\beta_1$, $\beta_2$, and $\beta_3$ are learnable task-balancing parameters. All losses are optimized jointly, so the segmentation decoder, class-response projection, BiFusion updates, and concept-aligned encoder representations are trained under the same segmentation supervision.

\section{Experiments}

\label{sec:experiments}

\begin{table*}[t]
\centering

\resizebox{\textwidth}{!}{
\begin{tabular}{l|ccc|ccc|ccc|ccc}
\toprule
        \multirow{3}{*}{Method}&\multicolumn{3}{c|}{AMOS22 (CT)}&\multicolumn{3}{c|}{MM-WHS (CT)}&\multicolumn{3}{c|}{MM-WHS (MR)}&\multicolumn{3}{c}{MSD-Brain (MR)}\\
        \cline{2-13}
        &DSC$\uparrow$&HD95$\downarrow$&NSD$\uparrow$&DSC$\uparrow$&HD95$\downarrow$&NSD$\uparrow$&DSC$\uparrow$&HD95$\downarrow$&NSD$\uparrow$&DSC$\uparrow$&HD95$\downarrow$&NSD$\uparrow$\\
        \hline
        \hline 
         U-Net~\cite{ronneberger2015unet} & 80.64 & 8.88 & 87.99 & 89.19&3.94&93.62&76.33&\underline{28.44}&76.61&68.73 & 15.70 & 73.33\\
        UNet++~\cite{zhou2018unet++}& 84.30 & 11.46 & 91.14 & 90.29&4.14&93.91&79.37&42.85&78.42& 74.10 & 12.05 & 76.98 \\
        nnU-Net~\cite{isensee2021nnu}& \underline{86.33} & 9.05 & 92.01 & 90.05&4.03&93.72&\underline{80.71}&46.37&79.21& 75.10 & 12.38 & 78.29\\
        STU-Net~\cite{huang2023stu}& 84.21 & 8.88 & 91.42 & 90.85&3.88&94.22&77.11&37.51&77.25&72.05&11.95&77.97\\
        MedNeXt~\cite{roy2023mednext}& 84.88 & 8.32 & 91.92 & 90.39&3.62&94.11&77.54&36.51&77.11&71.84&14.59&76.90\\
        \hline
        UNETR ~\cite{hatamizadeh2022unetr} & 77.67 & 12.20 & 83.22 & 88.96&4.09&91.84&74.72&30.46&75.44& 70.78 & 14.25 & 75.11\\
        Swin UNETR ~\cite{hatamizadeh2021swin} & 84.71 & 10.84 & 90.74 & 90.48&3.83&93.83&80.46&37.37&80.40&\textbf{75.82} & 12.31 & 78.76\\
        SwinUNETR-V2 ~\cite{he2023swinunetr} & 85.53 & 9.28 & 91.75
        & 90.51&3.85&94.34&80.04&35.52&\underline{81.08}&75.02 & 12.70 & 77.63\\
        nnFormer~\cite{zhou2021nnformer}& 76.98 & 10.98 & 84.57 & \underline{91.43}&\underline{3.52}&\underline{95.11}&74.54&36.75&77.03&72.44&12.83&77.55\\
        MedFormer~\cite{gao2022data}& 85.84 & 8.83 & \underline{92.03} & 90.68&3.79&93.83&78.18&37.88&78.64&74.64&\underline{10.32}&78.20\\
        
        \hline
        \hline
        Universal-CLIP~\cite{liu2023clip} & 82.59 & 11.25 & 88.02 & 90.14&4.23&93.27&77.07&44.30&76.71&72.31&12.65&75.85\\
        MulModSeg~\cite{li2024novel}& 83.79 & \underline{8.28} & 89.56 & 89.95&4.08&93.41&80.23&34.54&80.46&72.65&12.24&76.54\\
        ZePT~\cite{jiang2024zept}& 81.12 & 13.02 & 85.35 & 89.96&4.94&93.17&77.56&33.01&78.25&73.47&14.88&\underline{79.28}\\
        CAT~\cite{huang2024cat}& 83.35 & 9.56 & 88.55 & 90.22&3.98&93.11&77.01&38.17&79.25&71.14&15.64&77.19\\
        MedPlex (Ours)  & \textbf{88.21} & \textbf{6.52} & \textbf{93.97} & \textbf{92.60}&\textbf{3.09}&\textbf{96.86}&\textbf{82.94}&\textbf{27.01}&\textbf{83.14}&\underline{75.42}&\textbf{9.82}&\textbf{79.87}\\
        \bottomrule
    \end{tabular}
    }
    \caption{\small Comparison of MedPlex with benchmark vision-only models and VLMs for medical image segmentation. Vision-only models appear first, followed by VLMs (separated by a double horizontal line). The best results are \textbf{bolded} and the second-best results are \underline{underlined}.} 
    
    \label{tab:results}
\end{table*}

\begin{figure*}[t]
\centering
\includegraphics[width=1.0\textwidth]{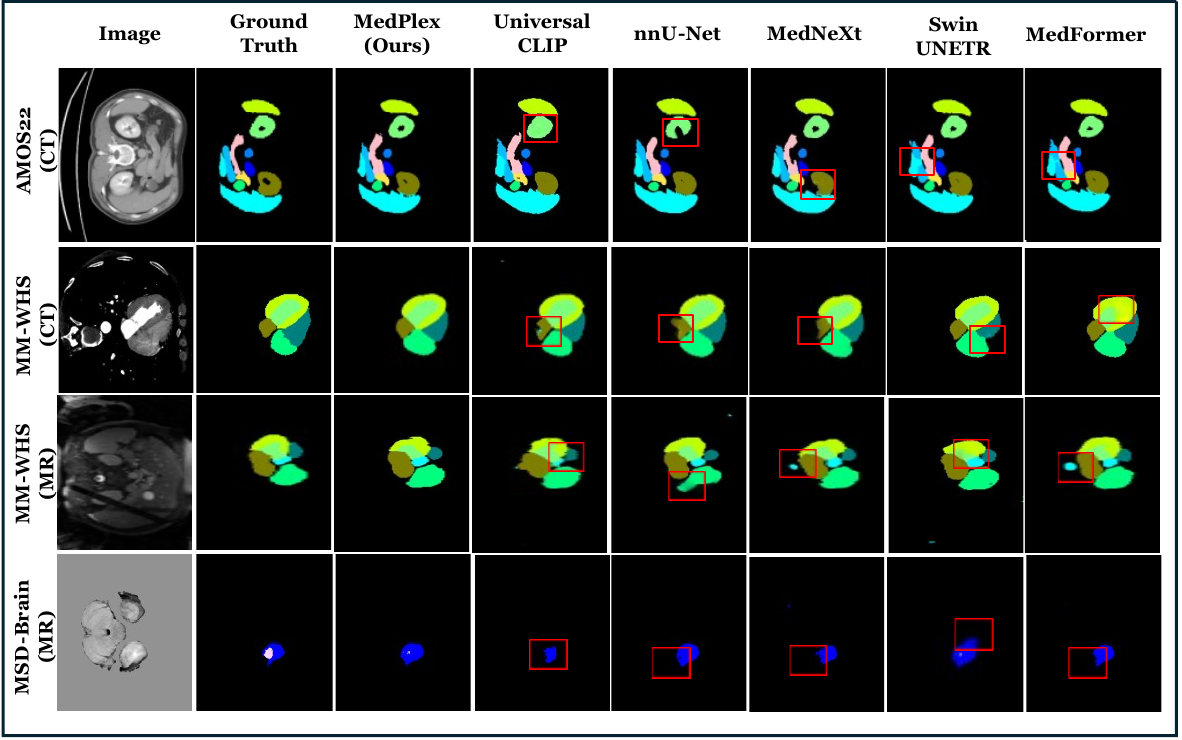}
\caption{\small Qualitative visualizations of randomly selected image slices from AMOS22, MM-WHS (CT), MM-WHS (MR), and MSD-Brain (MR). Comparison of MedPlex against other benchmark models, with red boxes indicating areas where MedPlex outperforms.}
\label{fig:figure5}
\vskip -0.1 in
\end{figure*}

\begin{figure}[t]
    \centering
    \includegraphics[width=0.7\columnwidth]{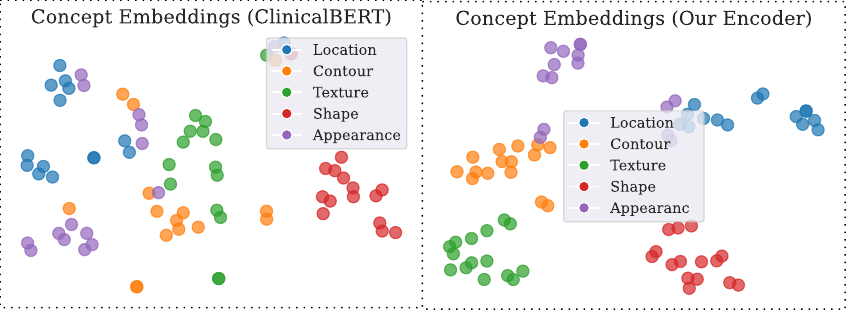}
    \caption{\small t-SNE visualization shows improved concept separation in MedPlex’s text encoder (right) compared to ClinicalBERT (left), demonstrating the benefit of concept-grounded contrastive alignment. \label{fig:figure6}}
    \vspace{-18pt}
\end{figure}

\subsection{Implementation Details}
\label{evaluation}

\noindent\textbf{Datasets.}
We evaluate MedPlex on 3D CT and MR datasets covering diverse anatomical targets: AMOS22 (CT)~\cite{ji2022amos} and BTCV (CT)~\cite{landman2015miccai} for abdominal organs, MM-WHS (CT/MR)~\cite{zhuang2019evaluation} for cardiac substructures, and MSD-Brain (MR)~\cite{antonelli2022medical} for brain tumors. To evaluate real grounded clinical text, we use the focal abnormalities subset of ReXGroundingCT~\cite{baharoon2025ReXGroundingCT}, derived from CT-RATE~\cite{hamamci2024developing}, which pairs 3D CT images with free-text clinical reports grounded to voxel-level annotations. For datasets without report-level supervision, we pair each anatomical class with structured clinical descriptions generated from clinically observable concepts. Additional dataset details, preprocessing steps, and hyperparameter settings are provided in the Appendix.

\noindent\textbf{Experiment Setup.}
We initialize the text encoder with ClinicalBERT~\cite{alsentzer2019publicly} using Hugging Face Transformers~\cite{wolf2019huggingface} and use Swin UNETR~\cite{hatamizadeh2021swin} as the vision model. Models are trained for $300$ epochs with validation every $5$ epochs using AdamW with learning rate $10^{-4}$, weight decay $10^{-5}$, and a cosine annealing scheduler decaying to $10^{-7}$. All experiments are conducted on an NVIDIA GeForce RTX 4090 (24 GB) using Python $3.9.18$. Further hyperparameter details are provided in the Appendix.

\noindent\textbf{Baselines.}
We compare MedPlex against representative CNN-based, transformer-based, and vision-language segmentation models. All models are trained using the same data splits and optimization schedule. For a controlled comparison, all vision-language baselines receive the same underlying textual information, adapted to each method’s required input format, and follow the same training schedule as MedPlex.

\noindent\textbf{Evaluation Metrics.}
Performance is measured using Dice Similarity Coefficient (DSC, \%), 95\% Hausdorff Distance (HD95, mm), and Normalized Surface Distance (NSD, \%), which measures boundary agreement within a predefined tolerance. Higher DSC and NSD indicate better segmentation, while lower HD95 indicates smaller boundary error.


\subsection{Results and Discussion}

The experiments evaluate the central claims of MedPlex: whether continuous vision-language co-adaptation improves segmentation across CT and MR benchmarks, whether concept-grounded alignment makes textual supervision more effective, and whether the learned representations generalize across datasets, backbones, and language encoders.

\noindent\textbf{Main Segmentation Performance.}
\Cref{fig:figure5} presents qualitative comparisons between MedPlex, top-performing vision-only models, and vision-language baselines. Each row shows a randomly selected test slice with its ground truth and corresponding segmentation outputs across AMOS22 (CT), MM-WHS (CT/MR), and MSD-Brain (MR). MedPlex produces predictions closer to the ground truth, with red boxes highlighting low-contrast regions and ambiguous boundaries where it improves over competing models.

\begin{wraptable}{r}{0.45\textwidth}
\centering
\small
\resizebox{0.45\textwidth}{!}{
    \begin{tabular}{c|c|c}
        \toprule
        {\multirow{2}{*}{Method}} & \multicolumn{2}{c}{ReXGroundingCT} \\
        \cline{2-3}
        & DSC$\uparrow$ & NSD$\uparrow$ \\
        \hline \hline
        Universal-CLIP~\cite{liu2023clip}         & 18.99  & 19.98 \\
        MulModSeg~\cite{li2024novel}       & 19.05  & 23.45 \\
        ZePT~\cite{jiang2024zept}       & 17.45  & 22.05 \\
        CAT~\cite{huang2024cat}       & 18.78  & 28.64 \\
        MedPlex w/o BiFusion& 19.55 & 25.06 \\
        MedPlex w/o Alignment& 20.50  & 29.59 \\
        MedPlex (full)& \textbf{21.78}  & \textbf{31.32} \\
        \bottomrule
    \end{tabular}
}
\caption{\small 
Comparison under real clinical-text supervision on ReXGroundingCT.
}
\label{tab:real_text}
\vspace{-2mm}
\end{wraptable}

Quantitative results in \Cref{tab:results} further show that MedPlex improves over strong vision-only baselines, including CNN-based models, transformer-based models, and hybrid architectures such as MedFormer. On AMOS22, MedPlex surpasses the best vision-only result on each metric, improving DSC by nearly 2\%, reducing HD95 from 8.32 mm to 6.52 mm, and improving NSD from 92.03\% to 93.97\%. On MM-WHS (CT), it improves over the strongest vision-only baseline across all metrics, increasing DSC from 91.43\% to 92.60\%, reducing HD95 from 3.52 mm to 3.09 mm, and improving NSD from 95.11\% to 96.86\%. On MM-WHS (MR), MedPlex also achieves the best results across all metrics, improving DSC from 80.71\% to 82.94\%, HD95 from 28.44 mm to 27.01 mm, and NSD from 81.08\% to 83.14\%. On MSD-Brain, MedPlex achieves the best HD95 and NSD while remaining competitive in DSC. These results suggest that integrating textual descriptions into representation learning improves both region overlap and boundary quality across diverse anatomical targets and imaging modalities.

\noindent\textbf{Comparison with Text-Guided VLM Segmentation Methods.}
MedPlex outperforms prior 
\begin{wraptable}{r}{0.70\textwidth}
\small
\resizebox{0.70\textwidth}{!}{
    \begin{tabular}{ccc|c|c|c}
        \toprule
        \multicolumn{3}{c|}{\multirow{2}{*}{Textual Information Format}}& \multirow{2}{*}{Alignment}&\multicolumn{2}{c}{AMOS22 (CT)}\\
        \cline{5-6}
        &&&&DSC$\uparrow$&NSD$\uparrow$\\
        \hline \hline
\multicolumn{3}{c|}{Fixed, consistent concepts}  & Class \& Region & \textbf{88.21} & \textbf{93.97}\\
\multicolumn{3}{c|}{Non-fixed, ambiguous concepts} & Class \& Region & 87.24 & 93.22\\
\multicolumn{3}{c|}{Single aggregated definition}  & Class-level & 87.01 & 92.91\\
\multicolumn{3}{c|}{Class names, no concepts}      & Class-level & 86.45 & 91.34\\
        \bottomrule
    \end{tabular}
}
\caption{\small Effects of textual information format on MedPlex.}
\label{tab:ablation1}
\vspace{-2mm}
\end{wraptable}
VLM-based segmentation methods across datasets and modalities in \Cref{tab:results}, including approximately 4\% gains in both DSC and NSD on AMOS22. Qualitative results in \Cref{fig:figure5} also show improved segmentation of visually ambiguous regions compared with Universal-CLIP~\cite{liu2023clip}. 
These results indicate that encoder-level vision-language co-adaptation turns clinical concepts into class-specific evidence, improving the separation of structures with similar intensities, weak boundaries, and modality-dependent appearance.

On ReXGroundingCT (\Cref{tab:real_text}), supervision is closer to real clinical use: free-text reports describe focal abnormalities, and the corresponding masks are sparse rather than exhaustive. This setting is challenging because the text does not provide clean class-level descriptions, and the visual targets may be incomplete or weakly localized. MedPlex achieves the best DSC and NSD among all compared methods, showing that its advantage extends beyond controlled structured descriptions. The variants without BiFusion or alignment further clarify where this robustness comes from: removing concept-grounded organization lowers DSC to 20.50, while removing continuous grounding of textual representations in image-specific visual evidence lowers it further to 19.55, indicating that co-adaptation matters most when the language itself is noisy. These results suggest that MedPlex is better suited to realistic clinical-text supervision because it combines adaptive vision-language interaction with structured alignment, allowing noisy report-derived language to contribute to dense prediction more effectively.

\noindent\textbf{Effects of Textual Information.}
To analyze how the structure of textual supervision affects multimodal alignment, we evaluate MedPlex on the AMOS22 (CT) dataset in \Cref{tab:ablation1}. We compare multiple textual input formats: fixed, consistent concepts (default), non-fixed or ambiguous concepts without manual refinement, a single detailed sentence aggregating all concepts, and class names only.
\begin{wraptable}{r}{0.5\textwidth}
\centering
\small
\resizebox{0.5\textwidth}{!}{
    \begin{tabular}{c|c|c|c}
        \toprule
        {\multirow{2}{*}{Method}} & \multicolumn{3}{c}{AMOS22 (CT) $\rightarrow$ BTCV (CT)} \\
        \cline{2-4}
        & DSC$\uparrow$ & HD95$\downarrow$ & NSD$\uparrow$ \\
        \hline \hline
        nnU-Net~\cite{isensee2021nnu}         & 73.54 & 19.00 & 81.04 \\
        MedFormer~\cite{gao2022data}       & 72.84 & 17.06 & 80.06 \\
        Universal-CLIP~\cite{liu2023clip}       & 70.04 & 23.28 & 78.05 \\
        MedPlex (Ours)& \textbf{79.65} & \textbf{9.74} & \textbf{88.02} \\
        \bottomrule
    \end{tabular}
}
\vspace{-2mm}
\caption{\small 
Cross-benchmark transfer results on BTCV using models trained on AMOS22. Evaluation is performed without fine-tuning on BTCV.
}
\vspace{-4mm}
\label{tab:generalization}
\end{wraptable}

\noindent Consistent concept descriptions achieve the best performance, while ambiguous concepts reduce accuracy. Aggregating all concepts into a single description lowers DSC by 1.2 points, and using only class names produces a 1.8-point drop, showing that MedPlex benefits from explicit concept structure, not textual content alone. Removing all textual input (\Cref{tab:ablation2}) decreases DSC by 3.5 points, further confirming the value of structured semantic supervision.

\noindent\textbf{Cross-Benchmark Transfer.}
Models trained on AMOS22 are evaluated directly on BTCV without adaptation. As shown in \Cref{tab:generalization}, MedPlex achieves the best results, improving DSC from 73.54\% to 79.65\%, reducing HD95 from 17.06 mm to 9.74 mm, and increasing NSD from 81.04\% to 88.02\%. The substantial HD95 reduction indicates stronger boundary transfer, showing that MedPlex learns anatomical evidence that generalizes beyond the source benchmark.

\begin{wraptable}{r}{0.7\textwidth}
\small
\centering
\resizebox{0.7\textwidth}{!}{
\begin{tabular}{cccc|cc}
\toprule
\multicolumn{4}{c|}{MedPlex Components} &
\multicolumn{2}{c}{AMOS22 (CT)}\\
\hline
\multirow{2}{*}{Backbone} &
\multirow{2}{*}{Fusion} &
Class-Level &
Region-Level &
\multirow{2}{*}{DSC$\uparrow$} &
\multirow{2}{*}{NSD$\uparrow$}\\
& & Alignment & Alignment & & \\
\hline\hline
UNet++     & BiFusion & $\checkmark$ & $\checkmark$ & 86.47 & 92.52 \\
Swin UNETR & BiFusion & $\checkmark$ & $\checkmark$ & \textbf{88.21} & \textbf{93.97} \\
\hline
Swin UNETR & BiFusion & $\checkmark$ & $\times$     & 87.54 & 92.89 \\
Swin UNETR & BiFusion & $\times$     & $\times$     & 86.71 & 92.07 \\
Swin UNETR & BiFusion & $\times$     & $\checkmark$ & 86.87 & 92.24 \\
\hline
Swin UNETR & Unidirectional & $\checkmark$ & $\checkmark$ & 85.98 & 92.27 \\
Swin UNETR & Late Fusion    & $\checkmark$ & $\checkmark$ & 85.56 & 91.66 \\
Swin UNETR & None           & $\times$     & $\times$     & 84.71 & 90.74 \\
\bottomrule
\end{tabular}
}
\caption{\small Ablation of MedPlex components and fusion designs.}
\vspace{-4mm}
\label{tab:ablation2}
\end{wraptable}

\noindent\textbf{Component Analysis.}
\Cref{tab:ablation2} evaluates the contribution of each MedPlex component. With all components enabled, both backbones improve over their vision-only counterparts in \Cref{tab:results}, showing that the framework is not tied to a single backbone. Removing region-level concept alignment, both alignment objectives, or class-level concept alignment consistently reduces performance. Region-level concept alignment is most effective when supported by class-level anatomical anchors.
The matched unidirectional fusion strategy updates the vision encoder with text across the same stages but does not update textual representations using visual evidence. It reaches 85.98 DSC and 92.27 NSD, compared with 88.21 DSC and 93.97 NSD for BiFusion, while late fusion obtains 85.56 DSC and 91.66 NSD. These comparisons isolate the benefit of bidirectional co-adaptation from multi-stage text injection. Removing fusion and alignment produces the lowest overall performance.

\noindent\textbf{Representation Analysis.}
Beyond segmentation accuracy, we examine whether concept-
\begin{wraptable}{r}{0.45\textwidth}
\vspace{-4mm}
\small
\centering
\resizebox{0.44\textwidth}{!}{
\begin{tabular}{cc|cc}
\toprule
\multicolumn{2}{c|}{MedPlex} & \multicolumn{2}{c}{AMOS22 (CT)} \\
\hline
Vision & Language & DSC$\uparrow$ & NSD$\uparrow$ \\
\hline\hline
Swin UNETR & ClinicalBERT & 88.21 & 93.97 \\
Swin UNETR & BioBERT      & 87.91 & 93.19 \\
Swin UNETR & PubMedBERT   & 88.01 & 93.82 \\
\hline
U-Net        & ClinicalBERT & 83.79 & 90.36 \\
UNETR       & ClinicalBERT & 79.88 & 87.45 \\
\bottomrule
\end{tabular}
}
\caption{\small Ablation on vision and language backbone choices.}
\label{tab:ablation3}
\vspace{-4mm}
\end{wraptable}
grounded alignment changes the structure of the textual representation space. The t-SNE plots in \Cref{fig:figure6} compare ClinicalBERT with the MedPlex text encoder after training. ClinicalBERT separates coarse concepts but shows weaker organization for domain-specific anatomical semantics, whereas MedPlex produces clearer separation of fine-grained and anatomically meaningful concepts. This supports the role of concept-grounded contrastive alignment in preserving clinical concept structure, which is central to using textual descriptions as segmentation supervision.

\begin{wraptable}{r}{0.52\textwidth}
\vspace{-4mm}
\small
\centering
\resizebox{0.51\textwidth}{!}{
\begin{tabular}{c|ccccc}
\toprule
Method & Params & FLOPs & Infer. & Train & Iter \\
       & (M)    & (GMac) & Mem (GB) & Mem (GB) & Time (s) \\
\hline\hline
Vision-Only  & 62.19  & 328.94 & 1.54 & 3.15 & 8.53  \\
ZePT         & 189.77 & 360.50 & 2.43 & 7.54 & 14.47 \\
MedPlex   & 183.46 & 357.26 & 2.36 & 7.33 & 13.89 \\
\bottomrule
\end{tabular}
}
\caption{\small Computational profiles of MedPlex, ZePT, and vision-only baseline.}
\label{tab:cost_analysis}
\vspace{-2mm}
\end{wraptable}

\noindent\textbf{Effect of Language Backbone.}
Table~\ref{tab:ablation3} shows that replacing ClinicalBERT with BioBERT or PubMedBERT leads to only minor performance changes, suggesting that MedPlex is not highly sensitive to the initial language encoder. This indicates that the hierarchical fusion and concept-grounded alignment objectives help adapt different biomedical text encoders into a shared vision-language representation space. Replacing the vision backbone (Table~\ref{tab:ablation3}) with U-Net or UNETR also yields consistent gains over their vision-only counterparts from \Cref{tab:results}, showing that the benefits of BiFusion and concept-grounded contrastive alignment are not tied to a specific segmentation architecture.

\noindent\textbf{Computational Analysis.}
As shown in Table~\ref{tab:cost_analysis}, MedPlex introduces modest overhead relative to a vision-only baseline due to multimodal integration. Compared with ZePT~\cite{jiang2024zept}, MedPlex uses fewer parameters (183M vs. 190M), lower GPU memory during both training and inference, and slightly faster per-iteration training time (13.89s vs. 14.47s). These results indicate that the BiFusion and alignment modules add limited computational cost while improving segmentation performance, giving MedPlex a favorable efficiency-performance tradeoff.

\section{Conclusion}
\label{sec:conclusion}

We present MedPlex, a vision-language framework that turns textual guidance into continuous, clinically grounded representation learning through BiFusion and concept-grounded alignment. Across CT and MR benchmarks, including free-text report supervision, MedPlex consistently improves segmentation across diverse anatomical targets and textual settings.

\noindent\textbf{Limitations.}
MedPlex uses volume- or anatomical-class-level textual supervision without explicitly modeling 2D slice-wise cues. Future work will explore slice-aware supervision for localized clinical descriptions and more lightweight model variants.

\bibliography{egbib}

\appendix

\section{Appendix: Additional Technical Details and Benchmarks}
\label{sec:appendix_details}

\subsection{Additional Method Details}
\label{sec:appendix_method}

\noindent\textbf{Vision Model.}
MedPlex can be integrated with standard encoder-decoder 3D segmentation backbones. In our main experiments, we use Swin UNETR~\cite{hatamizadeh2021swin} as the vision model. Given a 3D medical image $\mathbf{X} \in \mathbb{R}^{H \times W \times D \times 1}$, the vision encoder produces multi-stage visual representations that are used for stage-wise fusion with textual representations. The encoder is organized into four stages, each composed of Swin Transformer blocks~\cite{hatamizadeh2021swin} with MLP layers, layer normalization, residual connections, and downsampling between stages. At stage $n$, the visual feature map has spatial resolution approximately $(H/2^n, W/2^n, D/2^n)$.

The decoder receives the BiFusion-updated encoder representations through skip connections at the corresponding resolutions. These skip connections combine fine-grained spatial detail from earlier stages with higher-level semantic representations from deeper stages. The final prediction layer maps the decoder output to an $N$-class segmentation map, where $N$ denotes the number of anatomical classes.

\noindent\textbf{Text Encoder.}
We evaluate MedPlex with three domain-specific BERT encoders: BioBERT~\cite{lee2020biobert}, ClinicalBERT~\cite{alsentzer2019publicly}, and PubMedBERT~\cite{gu2021domain}. BioBERT is pretrained on biomedical corpora such as PubMed abstracts and PMC full-text articles, ClinicalBERT is adapted to clinical notes and electronic health records, and PubMedBERT is trained from scratch on PubMed abstracts. These models provide biomedical or clinical textual representations that can be incorporated into MedPlex through the stage-wise language stream. In the main experiments, we use ClinicalBERT as the default text encoder and report backbone sensitivity results in the main paper.

\noindent\textbf{Segmentation Loss.}
The segmentation objective combines Dice loss and cross-entropy loss:
\begin{equation}
\mathcal{L}_{\mathrm{seg}}
=
\mathcal{L}_{\mathrm{dice}}(\mathbf{P}, \mathbf{G})
+
\mathcal{L}_{\mathrm{ce}}(\mathbf{P}, \mathbf{G}),
\end{equation}
where $\mathbf{P}$ denotes the predicted segmentation map and $\mathbf{G}$ denotes the ground-truth mask. The Dice loss is defined as
\begin{equation}
\mathcal{L}_{\mathrm{dice}}(\mathbf{P}, \mathbf{G})
=
1 -
\frac{
2\sum_j \mathbf{P}_j \mathbf{G}_j
}{
\sum_j \mathbf{P}_j^2 + \sum_j \mathbf{G}_j^2 + \epsilon
},
\end{equation}
where $j$ indexes voxels and $\epsilon$ is a small constant for numerical stability. The cross-entropy term provides voxel-wise class supervision and complements the overlap-based Dice objective.

\noindent\textbf{Balancing Multiple Loss Functions.}
MedPlex jointly optimizes segmentation, class-response supervision, and concept-grounded alignment:
\begin{equation}
\mathcal{L}
=
\beta_1\mathcal{L}_{\mathrm{seg}}
+
\beta_2\mathcal{L}_{\mathrm{ClassEmbed}}
+
\beta_3\mathcal{L}_{\mathrm{alignment}}.
\end{equation}
Following homoscedastic uncertainty weighting~\cite{kendall2018multi}, we parameterize each learnable task weight as
$\beta_i = 1/(2\sigma_i^2)$. Including the corresponding regularization terms, the complete training objective is

\begin{align}
\mathcal{L}
&=
\frac{1}{2\sigma_1^2}\mathcal{L}_{\mathrm{seg}}
+
\frac{1}{2\sigma_2^2}\mathcal{L}_{\mathrm{ClassEmbed}}
+
\frac{1}{2\sigma_3^2}\mathcal{L}_{\mathrm{alignment}}
\notag\\
&\quad
+
\log\sigma_1
+
\log\sigma_2
+
\log\sigma_3,
\end{align}

where each $\sigma_i$ is learned during training and represents the uncertainty associated with one objective. Objectives with higher learned uncertainty receive lower weights, while the logarithmic terms prevent degenerate solutions. This formulation adaptively balances the three objectives without manually tuned static weights.

\subsection{Controlled Textual Information}
\label{sec:appendix_textual_information}

\noindent\textbf{Prompt for Structured Clinical Descriptions.}
For datasets without report-level supervision, we generate structured class-specific textual descriptions using an LLM. Each anatomical class is described through clinically observable concepts: shape, location, appearance/density, contour/symmetry, and internal texture. The following prompt template is used:

\begin{mdframed}[backgroundcolor=lightgray]
\small
\noindent ``[organ name] in a [modality] image.''

\noindent ``Shape: visual description of the shape of [organ name] in [modality] image.''

\noindent ``Location: visual description of the location of [organ name] in [modality] image.''

\noindent ``Appearance/Density: visual description of the appearance or density of [organ name] in [modality] image.''

\noindent ``Contour/Symmetry: visual description of the contour or symmetry of [organ name] in [modality] image.''

\noindent ``Internal Texture: visual description of the internal texture of [organ name] in [modality] image.''

\noindent Can you provide six total descriptions for each class in this format?

\noindent Do this for the following organs, and the modality is CT: left ventricle, right ventricle, left atrium, right atrium, myocardium of left ventricle, ascending aorta, and pulmonary artery.

\noindent We need visual descriptions only, not functional descriptions.
\end{mdframed}

\noindent\textbf{Concept Description Review Protocol.}
All generated concept descriptions were manually reviewed by the authors and cross-checked through a secondary LLM-based quality pass. The review verified that each description followed its requested concept category and contained only observable anatomical characteristics, without functional claims or clinical interpretation. The reviewed descriptions were then fixed and used consistently throughout training.

\Cref{tab:concepts_examples} shows representative examples of the class-specific structured descriptions used in our experiments.

\subsection{Datasets and Preprocessing}
\label{dataset_appendix}

\subsubsection{Dataset Description}
We evaluate MedPlex on diverse 3D medical imaging datasets covering abdominal organs, cardiac substructures, brain tumors, and thoracic abnormalities.

\noindent\textbf{AMOS22.} AMOS22~\cite{ji2022amos} provides abdominal CT images with annotations for 15 organs: spleen, right kidney, left kidney, gallbladder, esophagus, liver, stomach, aorta, inferior vena cava, pancreas, right adrenal gland, left adrenal gland, duodenum, bladder, and prostate/uterus. The dataset contains 200 training-validation images and 100 test images.

\noindent\textbf{BTCV.} BTCV~\cite{landman2015miccai} contains contrast-enhanced abdominal CT images annotated for 13 abdominal organs: spleen, right kidney, left kidney, gallbladder, esophagus, liver, stomach, aorta, inferior vena cava, portal vein and splenic vein, pancreas, right adrenal gland, and left adrenal gland.

\noindent\textbf{MM-WHS.} MM-WHS~\cite{zhuang2019evaluation} includes whole-heart segmentation from CT and MR images. It provides labels for seven cardiac structures: left ventricle blood cavity, myocardium of the left ventricle, right ventricle blood cavity, left atrium blood cavity, right atrium blood cavity, ascending aorta, and pulmonary artery. The dataset contains 20 CT and 20 MR images, which we evaluate separately as MM-WHS (CT) and MM-WHS (MR).

\noindent\textbf{MSD-Brain.} MSD-Brain~\cite{antonelli2022medical} is part of the Medical Segmentation Decathlon and focuses on brain tumor segmentation. Following common brain tumor segmentation protocols, we derive three tumor subregions: tumor core (TC), whole tumor (WT), and enhancing tumor (ET). Labels 2 and 3 are merged to form TC, labels 1, 2, and 3 are combined to form WT, and label 3 is used for ET. We use 484 publicly available cases, split into 349 training, 39 validation, and 96 test cases.

\noindent\textbf{ReXGroundingCT.}
ReXGroundingCT~\cite{baharoon2025ReXGroundingCT} is a 3D chest CT dataset that grounds free-text radiology findings to voxel-level segmentation masks. It is derived from CT-RATE~\cite{hamamci2024developing}, which pairs non-contrast chest CT volumes with clinical reports. We use the \emph{Focal Abnormalities} subset, containing localized pathological findings such as nodules, consolidations, and focal opacities. Our experimental split contains 1,680 training volumes and 264 test volumes. Unlike fixed-anatomy segmentation, each target must be identified from its free-text description. The reports were originally written in Turkish, machine-translated into English, and subsequently standardized using an LLM-based rewriting pipeline. Moreover, the training annotations contain at most three representative instances per finding, leaving additional visible instances unlabeled. These linguistic and annotation challenges make ReXGroundingCT a difficult benchmark for free-text medical image segmentation.

\subsubsection{Dataset Preprocessing}
We apply a standardized preprocessing pipeline to all 3D datasets. Images are reoriented to a consistent RAS coordinate system, resampled to dataset-specific voxel spacing, intensity-clipped when applicable, and normalized according to the modality. During training, we sample fixed-size 3D patches of size \(96 \times 96 \times 96\), guided by foreground labels to balance anatomical regions. Data augmentation includes random flipping along the three spatial axes, random rotations, and probabilistic intensity shifts to improve robustness to spatial and scanner variability. Dataset-specific preprocessing details are summarized in \Cref{tab:dataset_processing}.

\begin{table*}[t]
\centering
\small
\resizebox{1\linewidth}{!}{
\begin{tabular}{p{3.2cm}|p{3.3cm}|p{3.5cm}|p{3.7cm}|p{3.5cm}|p{3.5cm}}
\toprule
\textbf{Title} & \textbf{Shape} & \textbf{Location} & \textbf{Appearance/Density} & \textbf{Contour/Symmetry} & \textbf{Internal Texture} \\
\midrule
Spleen in a CT Image
& Ovoid or crescent-shaped with a smooth outer surface.
& Located in the left upper quadrant, lateral to the stomach and superior to the left kidney.
& Soft-tissue density with homogeneous enhancement after contrast administration.
& Well-defined smooth contour, slightly convex laterally and concave medially.
& Homogeneous fine-grained internal texture under normal conditions. \\
\midrule
Right Kidney in a CT Image
& Bean-shaped with a slightly concave medial border at the hilum.
& Positioned in the right retroperitoneum, inferior to the liver and lateral to the inferior vena cava.
& Intermediate soft-tissue density with a hypodense central collecting system.
& Smooth and well-defined outer contour with subtle lobulations in some cases.
& Cortex appears homogeneous, while the medulla may show striated texture due to renal pyramids. \\
\midrule
Left Kidney in a CT Image
& Bean-shaped with a slightly concave medial border at the hilum.
& Positioned in the left retroperitoneum, inferior to the spleen and lateral to the aorta.
& Intermediate soft-tissue density with a hypodense central collecting system.
& Smooth and well-defined outer contour with subtle lobulations in some cases.
& Cortex appears homogeneous, while the medulla may show striated texture due to renal pyramids. \\
\midrule
Left Ventricle in a CT Image
& Oval-like shape with thick walls tapering slightly toward the apex.
& Positioned on the left side of the heart, below the left atrium and anterior to the descending aorta.
& High-density muscular wall visible with contrast enhancement; the lumen appears lower in density when filled with blood or contrast agent.
& Smooth and well-defined contour with relatively symmetric walls around the cavity.
& Relatively homogeneous muscular structure, with myocardial striations sometimes visible. \\
\midrule
Right Ventricle in a CT Image
& Crescent-like shape with thinner walls compared with the left ventricle.
& Positioned on the right side of the heart, anterior and slightly medial to the left ventricle.
& Lower-density walls compared with the left ventricle, with the lumen often appearing darker due to blood or contrast agent.
& Less symmetric contour than the left ventricle, with variable wall thickness.
& Slightly heterogeneous texture due to trabeculae carneae along the inner surface. \\
\midrule
Left Atrium in a CT Image
& Elliptical or spherical chamber with relatively thin walls.
& Superior to the left ventricle and posterior to the right atrium.
& Thin, low-density walls surrounding a central lumen that appears darker due to blood or contrast.
& Smooth rounded contour with relatively uniform wall thickness.
& Homogeneous texture without prominent internal structures under normal conditions. \\
\bottomrule
\end{tabular}}
\caption{\small Examples of structured clinical concept descriptions used as textual inputs for concept-grounded alignment.}
\label{tab:concepts_examples}
\end{table*}

\begin{table}[t]
\centering
\small
\resizebox{1\columnwidth}{!}{
\begin{tabular}{l|c|c|c}
\toprule
\textbf{Dataset} & \textbf{Voxel Spacing (mm)} & \textbf{Intensity Range} & \textbf{Normalization} \\
\midrule
AMOS22 (CT) & \(1.5 \times 1.5 \times 2.0\) & \([-175, 250]\) HU & \([0,1]\) \\
BTCV (CT) & \(1.5 \times 1.5 \times 2.0\) & \([-175, 250]\) HU & \([0,1]\) \\
MM-WHS (CT) & \(1.5 \times 1.5 \times 2.0\) & \([0, 400]\) HU & \([0,1]\) \\
MM-WHS (MR) & \(1.5 \times 1.5 \times 2.0\) & Not applicable & Non-zero voxel normalization \\
MSD-Brain (MR) & \(1.0 \times 1.0 \times 1.0\) & Not applicable & Non-zero voxel normalization \\
ReXGroundingCT (CT) & \(1.0 \times 1.0 \times 1.0\) & \([-1000, 1000]\) HU & \([-1,1]\) \\
\bottomrule
\end{tabular}}
\caption{\small Dataset-specific preprocessing settings.}
\label{tab:dataset_processing}
\end{table}

\subsection{Hyperparameters}
\label{sec:appendix_hyperparameters}

\Cref{tab:parameter_tuning} summarizes the key hyperparameters used in our experiments.

\begin{table}[t]
\centering
\small
\resizebox{0.75\columnwidth}{!}{
\begin{tabular}{l|c}
\toprule
\textbf{Parameter} & \textbf{Value} \\
\midrule
$\alpha_1$, $\alpha_2$ & 1, 1 \\
$\beta_1$, $\beta_2$, $\beta_3$ & Learnable during training \\
Temperature $\tau$ & 0.2 \\
Swin UNETR base feature dimension & 48 \\
ClinicalBERT feature dimension & 768 \\
Patch size & \(96 \times 96 \times 96\) \\
Optimizer & AdamW \\
Learning rate & \(10^{-4}\) \\
Weight decay & \(10^{-5}\) \\
Scheduler minimum learning rate & \(10^{-7}\) \\
\bottomrule
\end{tabular}}
\caption{\small Key hyperparameters and model dimensions used in MedPlex.}
\label{tab:parameter_tuning}
\end{table}

\subsection{Evaluation Metrics}
\label{sec:appendix_metrics}

We evaluate segmentation performance using Dice Similarity Coefficient (DSC, \%), 95\% Hausdorff Distance (HD95, mm), and Normalized Surface Distance (NSD, \%)~\cite{nikolov2018deep}.

\noindent\textbf{Dice Similarity Coefficient.}
DSC measures overlap between the predicted segmentation map $\mathbf{P}$ and ground-truth mask $\mathbf{G}$:
\begin{equation}
\mathrm{DSC}
=
\frac{
2\sum_j \mathbf{P}_j \mathbf{G}_j
}{
\sum_j \mathbf{P}_j^2 + \sum_j \mathbf{G}_j^2 + \epsilon
}
\times 100,
\end{equation}
where $j$ indexes voxels and $\epsilon$ is a small constant for numerical stability. Higher DSC indicates better region overlap.

\noindent\textbf{95\% Hausdorff Distance.}
HD95 measures boundary discrepancy between the predicted and ground-truth surfaces while reducing sensitivity to extreme outliers. Let $\partial \mathbf{P}$ and $\partial \mathbf{G}$ denote the predicted and ground-truth surfaces. We define the bidirectional surface distance set as
\begin{equation}
\mathcal{D}
=
\left\{
\min_{\mathbf{y}\in\partial \mathbf{G}}\|\mathbf{x}-\mathbf{y}\|_2
:
\mathbf{x}\in\partial \mathbf{P}
\right\}
\cup
\left\{
\min_{\mathbf{x}\in\partial \mathbf{P}}\|\mathbf{y}-\mathbf{x}\|_2
:
\mathbf{y}\in\partial \mathbf{G}
\right\}.
\end{equation}
HD95 is then computed as
\begin{equation}
\mathrm{HD95}
=
\operatorname{percentile}_{95}(\mathcal{D}).
\end{equation}
Lower HD95 indicates better boundary alignment.

\noindent\textbf{Normalized Surface Distance.}
NSD measures the fraction of predicted and ground-truth surface points that lie within a predefined tolerance $\tau$ of the opposite surface. Let $d(\mathbf{x}, \partial \mathbf{G}) = \min_{\mathbf{y}\in\partial \mathbf{G}}\|\mathbf{x}-\mathbf{y}\|_2$ and $d(\mathbf{y}, \partial \mathbf{P}) = \min_{\mathbf{x}\in\partial \mathbf{P}}\|\mathbf{y}-\mathbf{x}\|_2$. NSD is defined as
\begin{equation}
\mathrm{NSD}
=
\frac{
\left|\{\mathbf{x}\in\partial \mathbf{P}: d(\mathbf{x},\partial \mathbf{G}) \leq \tau\}\right|
+
\left|\{\mathbf{y}\in\partial \mathbf{G}: d(\mathbf{y},\partial \mathbf{P}) \leq \tau\}\right|
}{
|\partial \mathbf{P}| + |\partial \mathbf{G}|
}
\times 100.
\end{equation}
Higher NSD indicates better surface agreement within the tolerance threshold.

\subsection{Benchmark Models}
\label{sec:appendix_benchmarks}

\noindent\textbf{Vision-Only Baselines.}
We compare MedPlex with representative CNN-based, transformer-based, and hybrid 3D segmentation models.

\begin{itemize}
    \item \textbf{U-Net}~\cite{ronneberger2015unet}: a CNN-based encoder-decoder architecture with skip connections for preserving spatial detail during upsampling.
    
    \item \textbf{UNet++}~\cite{zhou2018unet++}: an extension of U-Net with nested dense skip pathways for improved feature propagation.
    
    \item \textbf{nnU-Net}~\cite{isensee2021nnu}: a self-configuring segmentation framework that adapts preprocessing, architecture, and training settings to each dataset.
    
    \item \textbf{STU-Net}~\cite{huang2023stu}: a scalable supervised pretraining framework for medical image segmentation with strong transferability across datasets.
    
    \item \textbf{MedNeXt}~\cite{roy2023mednext}: a ConvNeXt-inspired medical segmentation model designed to combine convolutional inductive bias with scalable architecture design.
    
    \item \textbf{UNETR}~\cite{hatamizadeh2022unetr}: a transformer-based 3D segmentation model that uses a Vision Transformer encoder with a CNN-style decoder.
    
    \item \textbf{Swin UNETR}~\cite{hatamizadeh2021swin}: a hierarchical transformer-based encoder-decoder architecture using shifted-window self-attention.
    
    \item \textbf{SwinUNETR-V2}~\cite{he2023swinunetr}: an enhanced Swin UNETR variant with stage-wise convolutional improvements for stronger 3D segmentation.
    
    \item \textbf{nnFormer}~\cite{zhou2021nnformer}: a 3D transformer segmentation architecture designed to model long-range dependencies while preserving local structure.
    
    \item \textbf{MedFormer}~\cite{gao2022data}: a data-efficient transformer-based segmentation model using locality-sensitive attention to balance global context and efficiency.
\end{itemize}

\noindent\textbf{Text-Guided VLM Segmentation Baselines.}
We compare MedPlex with representative text-guided segmentation methods within the VLM paradigm.

\begin{itemize}
    \item \textbf{Universal-CLIP}~\cite{liu2023clip}: a CLIP-based text-guided segmentation approach that incorporates text embeddings for multi-organ and tumor segmentation. We use Swin UNETR~\cite{hatamizadeh2021swin} as the vision backbone and follow the original text-input format.
    
    \item \textbf{MulModSeg}~\cite{li2024novel}: a multimodal medical image segmentation framework that uses modality-conditioned text embeddings and alternating CT/MR training. We follow a setting comparable to Universal-CLIP due to their similar text-guided formulation.
    
    \item \textbf{ZePT}~\cite{jiang2024zept}: a zero-shot pan-tumor segmentation framework based on query disentanglement and self-prompting. For textual input, we aggregate our structured concept descriptions into a single description per class to match its prompt format.
    
    \item \textbf{CAT}~\cite{huang2024cat}: a dual-prompt segmentation framework that coordinates anatomical and textual prompts through a query-based design. We follow the same text-input setting used for ZePT because the two approaches use related prompt-based formulations.
\end{itemize}

\end{document}